\documentclass[letterpaper, 10 pt, conference]{ieeeconf}  %

\usepackage{amsmath,amsfonts,amssymb}
\usepackage{mathtools}
\usepackage{bbm}
\usepackage{epsfig} %
\usepackage{times} %
\usepackage[english]{babel}
\usepackage{graphicx}
\usepackage{subfigure}
\usepackage{nicefrac}

\newcommand{\BX}{\mathbb{X}}
\newcommand{\BY}{\mathbb{Y}}

\newcommand{\BS}{\mathbb{S}}
\newcommand{\BA}{\mathbb{A}}

\newcommand{\BI}{\mathbb{I}}

\newcommand{\CF}{\mathcal{F}}

\newcommand{\CH}{\mathcal{H}}

\newcommand{\CR}{\mathcal{R}}

\def \CF{\mathcal{F}}

\def \CD{\mathcal{D}}

\newcommand{\tr}{^\mathrm{T}}  %
\newcommand{\RR}{\mathbb{R}}
\newcommand{\R}{\mathbb{R}}

\newcommand{\EE}{\mathbb{E}}
\newcommand{\PP}{\mathbb{P}}
\newcommand{\Var}{\mbox{Var}}

\DeclareMathOperator {\sign}{sign}

\newtheorem{theorem}{Theorem}
\newtheorem{lemma}{Lemma}

\newcommand{\norm}[1]{\left\lVert#1\right\rVert}

\newcommand{\startsquarepar}{%
    \par\begingroup \parfillskip 0pt \relax}
\newcommand{\stopsquarepar}{%
    \par\endgroup}

\IEEEoverridecommandlockouts                              %

\title{\LARGE \bf
Semi-Parametric Uncertainty Bounds for Binary Classification
}
\author{Bal\'azs Csan\'ad Cs\'aji$^{1}$ \and Ambrus Tam\'as$^{1}$%
\thanks{*This work was supported by the National Research, 
Dev.\ and Innovation Office (NKFIH), Hungary, grant numbers ED\_18-2-2018-0006 and KH\_17 125698. B.~Cs.~Cs\'aji was supported by a J\'anos Bolyai Res.\ Fellowship.}%
\thanks{$^{1}$
Bal\'azs Csan\'ad Cs\'aji and Ambrus Tam\'as
are with MTA SZTAKI: The Institute for Computer Science and Control, Hungarian Academy of Sciences, Budapest, Hungary, {\tt\small  balazs.csaji@sztaki.mta.hu, tambrus96@gmail.com}}%
}

\begin{document}

\maketitle
\thispagestyle{empty}
\pagestyle{empty}

\begin{abstract}

The paper studies binary classification and aims at estimating the underlying regression function which is the conditional expectation of the class labels given the inputs. The regression function is the key component of the Bayes optimal classifier, moreover, besides providing optimal predictions, it can also assess the risk of misclassification. We aim at building non-asymptotic confidence regions for the regression function and suggest three kernel-based semi-parametric resampling methods.  We prove that all of them guarantee regions with exact coverage probabilities and they are strongly consistent.

\end{abstract}

\medskip
\section{Introduction}
Classification is one of the 
principal problems of {\em statistical learning theory} \cite{Vapnik1998}, and it is widely applied across several fields \cite{hofmann2008kernel}, 
for example, in quantized identification \cite{goudjil2015identification}. A typical aim of classification is to minimize the {\em probability of misclassification}. If the (joint) probability distribution of the input-output pairs was known, the misclassification probability could be minimized by the Bayes optimal classifier. This classifier can be written as the sign of the {\em regression} function which is the conditional expectation of the labels given the inputs. The regression function can also help to assess the {risk of misclassification}. Estimating the regression function can be seen as identifying a (nonlinear) function from a sample of input and {\em quantized} (binary) output measurements. 

Besides providing point-estimates of the regression function, for which there are several methods available \cite{Vapnik1998, gyorfi2002distribution}, it is also an important problem to bound the {\em uncertainty} of a candidate model. We will provide these bounds in the form of {\em confidence regions}. Note that such regions also induce confidence sets for the misclassification probabilities.

In this paper, inspired by recent developments in Finite-Sample System Identification (FSID) 
\cite{Algo2018, SPSPaper2ITA, KolumbanThesis2016, pillonetto2018kernel}, we suggest three semi-parametric {\em kernel-based} \cite{hofmann2008kernel} resampling algorithms to build {\em non-asymptotic} confidence regions for the regression function of binary classification. We
prove that each of these algorithms provides confidence sets with {\em exact} coverage probabilities, and they are {\em strongly consistent}, that is any false model will be (almost surely) excluded from the confidence regions, as the sample size tends to infinity. As the suggested algorithms build on distribution-free results and work directly with the samples, the constructions are not restricted to models parametrized by finite dimensional vectors, but also allow infinite dimensional model classes.

\medskip
\section{Preliminaries}

\subsection{Binary Classification}

We are given an i.i.d. sample,
$\CD=\lbrace (x_i, y_i) \rbrace_{i=1}^n$ from an unknown joint distribution $P$ of the $(X, Y)$ random vector, where $x_i \in \BX$ is the input and $y_i \in \lbrace +1,-1\rbrace$ is the label of the $i$\,th observation. We call any $g: \BX \rightarrow \lbrace +1, -1 \rbrace$ function a {\em classifier}. The {\em Bayes optimal classifier} $g_*$ can be defined as the one which minimizes the a pirori risk functional $R(g) \doteq \mathbb{E}\big[\,L(Y, g(X)\,\big]$ where $L$ is an arbitrary loss function. 

In this paper we will focus on the $0 / 1$ loss that is one of the most common choices \cite{Vapnik1998}. It is 
defined by $L( y, g(x) ) \doteq\, \mathbb{I}\hspace{0.5mm}( g(x)\neq y)$, where 
$\mathbb{I}$ is the indicator function. 
The corresponding a priory risk is simply $R(g) \,=\, \mathbb{P}\hspace{0.5mm}(\,g(X)\neq Y\,)$. 

As distribution $P$ is unknown, 
we typically aim at estimating $g_*$.
At any point $x \in \BX$, $g_*(x) = \sign(\,\mathbb{E}\big[\,Y \,|\, X=x\,\big]\,)$ if it is feasible. Note that the conditional expectation $f_*(x) \doteq \mathbb{E}\big[\,Y \,|\, X=x\,\big]$ contains even more information than $g_*$, e.g., based on $f_*$ we are not only able to predict the label of a given input with minimal risk, but we can also calculate the risk itself, i.e., the probability of misclassification.
Therefore,  it is of high importance to study and estimate $f_*$.

\subsection{Reproducing Kernel Hilbert Spaces}

Given a Hilbert space $\mathcal{H}$ of $f: \BX \rightarrow \R$ type functions, with inner product $\langle\, \cdot, \cdot\, \rangle_\mathcal{H}$, we say that it is a {\em Reproducing Kernel Hilbert Space} (RKHS) if the point evaluation 
 function $\delta_x: f \rightarrow f(x)$ is bounded (or equivalently continuous) for all $x \in \BX$ \cite{hofmann2008kernel}. In this case, by the Riesz representation theorem, there uniquely exists $k(\cdot, \cdot)$, such that for all $x \in \BX$, $k(\cdot, x)  \in \mathcal{H}$ and  $f(x) = \langle\, f , k(\cdot,x) \,\rangle_\mathcal{H}$. This is called the {\em reproducing property}, and the function $k: \BX \times \BX \rightarrow \R$  is called the {\em kernel}. In particular $\langle \,k(\cdot,x) , k(\cdot,y) \,\rangle_\mathcal{H} =k(x,y)$ thus $k$ is symmetric and positive definite. The converse is also true by the Moore-Arnoszjan theorem \cite{aronszajn1950theory}: for each positive definite function there uniquely exists an RKHS. 
Typical examples of kernels are the Gaussian kernel, $k(x,y)= \exp(\frac{-\norm{x-y}^2}{2 \sigma^2})$ with $\sigma>0$, and the polynomial kernel, $k(x,y)=(x\tr y +c)^d$ with $c \geq 0$ and $d \in \mathbb{N}$.
For a given sample $\CD$, the Gram matrix, $K \in \RR^{n\times n}$, is defined as $K_{i,j}\! \doteq k(\,x_i, x_j\,)$, which is a (data-dependent) symmetric, positive semidefinite matrix. 

Let  $C_b(\BX)$ denote the space of bounded continuous functions on a compact metric space $\BX$. 
A kernel is {\em universal} if the corresponding $\mathcal{H}$ is dense in $C_b(\BX)$: 
for all $f \in C_b(\BX)$ and $\varepsilon > 0$ there exists $h \in \mathcal{H}$ such that $\norm{\hspace{0.3mm}f-h\hspace{0.3mm}}_\infty <\, \varepsilon$. 

\subsection{Kernel Mean Embedding}
The idea of {\em kernel mean embedding} is to map distributions to elements of an RKHS
with the help of the kernel \cite{muandet2017kernel}.
Let $(\BX, \Sigma)$ be a measurable space and let $M_{+}(\BX)$ denote the space of all probability measures on it. The kernel mean embedding of these probability measures into an RKHS $\mathcal{H}$ endowed with a reproducing kernel $k: \BX \times \BX \rightarrow \mathbb{R}$ is 
\begin{align}
\mu: M_{+}(\BX) \rightarrow \mathcal{H}, 
\\[1mm]
P \rightarrow \int k(x,\cdot)\, P(dx).
\end{align}

A kernel is called {\em characteristic} if the embedding, $\mu$, is {\em injective} (e.g., the Gaussian kernel). In this case the embedded element captures all informations about the distribution, e.g., 
for all $P,Q \in M_{+}(\BX)$,  $\norm{\hspace{0.3mm} \mu_P - \mu_Q\hspace{0.3mm} }_{\mathcal{H}} = 0$ if and only if $P=Q$. 
Hence, the embedding induces a metric on $M_{+}(\BX)$.

Let $\BX$ be a compact metric space and let $k$ be a universal kernel on $\BX$, then one can show that $k$ is also characteristic.

The kernel mean embedding has nice properties even when the 
kernel is not characteristic.
For example, for polynomial kernels with degree $d$ it holds that $\norm{\hspace{0.3mm} \mu_P - \mu_Q }_{\mathcal{H}\hspace{0.3mm}} = 0$ if and only if the first $d$ moments of $P$ and $Q$ are the same. 

Furthermore, many fundamental operations can be performed in $\mathcal{H}$ instead of dealing with the distributions themselves, e.g., Smola showed \cite{muandet2017kernel} that $\mathbb{E}_{P} [f(X)] = \langle f, \mu_{P} \rangle_{\mathcal{H}}$.

The underlying probability distribution of the sample is typically unknown, therefore, the kernel mean embedding should be estimated from empirical data. An important tool to prove the validity of such approaches is the
{\em Strong Law of Large Numbers} (SLLN) for random elements taking values in a {\em separable} Hilbert space $\CH$.
Let $\{X_n\}$ be a sequence of {\em independent} random elements taking values in $\CH$. If
\vspace{-0.7mm}
\begin{align}
\sum_{n=1}^\infty \frac{\Var(X_n)}{n^2}\, <\, \infty
\end{align}
where $\Var(X) \doteq\, \EE\big[\, \|\, X - \EE[X]\, \|_{\CH}^2\, \big]$, then
 \vspace{-0.7mm}
\begin{equation}
 \frac{1}{n}\, \sum_{k=1}^n (X_k -\mathbb{E} [X_k]) \, \rightarrow\, 0\quad\mbox{as}\quad n\to \infty,
\end{equation}
(a.s.) in the metric induced by $\|\cdot\|_{\CH}$ \cite[Theorem 3.1.4]{taylor1978stochastic}.

\medskip
\section{Resampling Framework}
\label{section:resampling-framework}
In this section we develop a {\em framework} to provide non-asymptotically guaranteed uncertainty quantification {\em resampling} algorithms for the ``regression function'', namely, the conditional expectation of the labels given the inputs. The regression function is a fundamental object to study, for example, its signs at various inputs define the Bayes optimal classifier which achieves minimal {\em misclassification risk}. 

Assume we have a (joint) distribution on\, $\BS\, \doteq\, \BX\, \times\, \BY$, where $\BX$ and $\BY$ are the input and output spaces, respectively. $\BX$ does not have to be $\RR^d$, but has to be a measurable space (with some $\sigma$-algebra). As we consider binary classification, \,$\BY \, \doteq \, \{+1,-1\}$.
The {\em regression function} can be written as
\begin{align}
\label{eq:regression-function}
f_*(x) \, \doteq\;& \, \EE \big[\,Y\; |\; X = x\, \big] \nonumber \\
=\;&\,  \PP(\,Y = +1\; |\; X = x\,) \,-\, \PP(\,Y = -1\; |\; X = x\,) \nonumber \\
=\;&\,  2 \cdot \PP(\,Y = +1\; |\; X = x\,) \, - \, 1.
\end{align}

Given $f_*$, the {\em Bayes optimal classifier} is
\begin{equation}
\label{eq:bayesian-classifier}
g_*(x) \, \doteq \; \sign(f_*(x)),
\end{equation}
where ``$\sign$'' denotes the signum function. Note that in \eqref{eq:bayesian-classifier}, for simplicity, we assumed that\, $\PP(\,f_*(X) \,\neq\, 0\,) \,=\, 1$.

We assume that we are given an (indexed) {\em family} of possible regression functions that also contains $f_*$, that is
\begin{equation}
f_* \in \CF \, \doteq \, \big\{\, f_{\theta}: \BX \to [\,-1,+1\,]\, \mid\, \theta \in \Theta\, \big\}.
\end{equation}

For simplicity, we refer to $\theta \in \Theta$ as a {\em parameter}, but $\Theta$ can be an {\em arbitrary} set, even an infinite dimensional vector space.
The true parameter is denoted by $\theta^*$, that is $f_{\theta^*} \,=\, f_*$. 

We assume that $\CF$ contains {\em square integrable} functions w.r.t.\ the input distribution, and that the parametrization is {\em injective}, i.e., $\theta_1 \,\neq\, \theta_2$ implies $f_{\theta_1} \,\neq\, f_{\theta_2}$ on a set having nonzero measure w.r.t.\ the input distribution. In other words,
\vspace{-2mm}
\begin{equation}
\|\, f_{\theta_1} - f_{\theta_2} \|^2_{\scriptscriptstyle P} \, \doteq \int_{\BX} (f_{\theta_1}\hspace{-0.4mm}(x)-f_{\theta_2}\hspace{-0.2mm}(x))^2 P_{\hspace{-0.7mm}\scriptscriptstyle\BX}(dx) \, \neq \, 0,
\label{eq:L2-distance-PX}
\end{equation}
if $\theta_1 \neq \theta_2$, where $P_{\hspace{-0.7mm}\scriptscriptstyle\BX}$ is the
distribution of the inputs.

Note that $f_*$ in itself does {\em not} determine the joint probability distribution generating the observations, namely, it does not contain information about the (marginal) distribution of the inputs, therefore, our approach is {\em semi-parametric}. 

As an example, consider the case where the ``$+1$'' class has probability density function $\varphi_1$, while the ``$-1$'' class has density $\varphi_2$. For each element of the sample, there is a $p$ probability to see an element with ``$+1$'' label and a $1-p$ probability to see a measurement with ``$-1$'' label. Then, 
\vspace{-0.7mm}
\begin{equation}
\EE \big[\,Y\;|\; X = x\, \big] \, =\; \frac{p\, \varphi_1(x) - (1-p) \,\varphi_2(x)}{p\, \varphi_1(x) + (1-p) \, \varphi_2(x)},
\label{eq:mixing-distributions}
\end{equation}
thus, if we have candidate densities for inputs with various labels and we know their mixing probability, then we can compute the regression function. However, observe that the regression function does not determine $\varphi_1, \varphi_2$ and $p$.

\subsection{Resampling Labels}
The observed i.i.d.\ input-output dataset is denoted by 
\begin{equation}
\CD_0 \,\doteq\, ((x_1, y_1), \dots, (x_n, y_n)),
\end{equation}
which can also be seen as a  $\BS^n$-valued random vector.

One of our core ideas
is that if we are given a candidate $\theta$, then we can generate (resample) alternative labels for the available inputs using the distribution induced by $f_{\theta}$, that is
\begin{align}
\label{eq:param-cond-prob}
\PP_{\theta}(\,Y = +1\, \mid\, X = x\,)\, =\; & \frac{f_{\theta}(x) + 1}{2},\nonumber\\
\PP_{\theta}(\,Y = -1\, \mid\, X = x\,)\, =\; & \frac{1 - f_{\theta}(x)}{2},
\end{align}
which immediatelly follow from our observations in \eqref{eq:regression-function}.
 
Given a $\theta$, we can generate $m-1$ {\em alternative samples} by
\begin{equation}
\CD_i(\theta) \,\doteq\, ((x_1, y_{i,1}(\theta)), \dots, (x_n, y_{i,n}(\theta))),
\end{equation}
for $i=1,\dots, m-1$, where for all $i,j$, label $y_{i,j}(\theta)$ is generated randomly according to the conditional distribution $\PP_{\theta}(\,Y \, \mid\, X = x_j\,)$. For notational simplicity, we extend this to $\CD_0$, that is
$\forall\,\theta:\CD_0(\theta) \doteq \CD_0$ and $\forall\,j:y_{0,j}(\theta)\, \doteq\, y_j$.

Naturally, for all $i$, dataset $\CD_i(\theta)$ can also be identified with a random vector in $\BS^n$, and $\CD_1(\theta), \dots, \CD_{m-1}(\theta)$ are always {\em conditionally i.i.d.}, given the inputs, $\{x_j\}$. 

Observe that, in case $\theta \neq \theta^*$, the distribution of $\CD_0$ is in general different than that of $\CD_i(\theta)$, $\forall \, i \neq 0$; while $\CD_0$ and $\CD_i(\theta^*)$  have the same distribution for all possible $i$.

\subsection{Ranking Functions}
The proposed algorithms will be defined via rank statistics based on suitably defined orderings. A key concept will be the ``ranking function'' which, informally, computes the rank of its first argument among all of its arguments based on some underlying ordering. Let 
$\BA$ be a measurable space
(with some $\sigma$-algebra), a (measurable) function 
$\psi : \BA^m \to [\,m\,]$, where $[\,m\,] \,\doteq\,\{1,\dots, m\}$, is called a 
{\em ranking function} if for all $(a_1, \dots, a_m) \in \BA^m$ it satisfies the two properties\medskip
\begin{enumerate}
\item[(P1)] For all permutations $\mu$ of the set $\{2,\dots, m\}$, 
we have 
\begin{equation*}
\psi\big(\,a_1, a_{2}, \dots, a_{m}\,\big)\; = \;
\psi\big(\,a_1, a_{\mu(2)}, \dots, a_{\mu(m)}\,\big),
\end{equation*}
that is the function is invariant with respect to reordering the last $m-1$ terms of its arguments.
\medskip
\item[(P2)] For all $i,j \in  [\,m\,]$,
if $a_i \neq a_j$, then we have
\begin{equation}
\psi\big(\,a_i, \{a_{k}\}_{k\neq i}\,\big)\, \neq \;\psi\big(\,a_j, \{a_{k}\}_{k\neq j}\,\big),
\end{equation}
where the simplified notation is justified by (P1).
\end{enumerate}
\medskip

We 
refer to the output of the ranking function $\psi$ as the {\em rank}. An important observation about ranking {\em exchangeable} \cite{kallenberg2002foundations} random elements is given by the following lemma. (Recall that if a sample is i.i.d., 
it is also exchangeable.)

\medskip
\begin{lemma}
\label{lemma:discrete-uniform-ranks}
{\em Let $A_1, \dots, A_m$ be exchangeable, almost surely pairwise different random elements taking values in $\BA$. Then,
$\psi\big(\,A_1, A_{2}, \dots, A_{m}\,\big)$ has discrete uniform distribution: for all $k \in [\,m\,]$, the rank is $k$ with probability $\nicefrac{1}{m}$.}
\end{lemma}
\medskip

\begin{proof}
Since $\{A_i\}$ are exchangeable, we know that
\begin{equation}
\label{eq:psi-exchangeable}
\PP \big(\,\psi\big(\,A_{1}, \dots, A_{m}\,\big)\,= \, k\,\big)\nonumber
\end{equation}
\begin{equation}
 = \;\PP \big(\,\psi\big(\,A_{\mu(1)}, \dots, A_{\mu(m)}\,\big)\, = \, k\,\big) ,
\end{equation}
for all $k\in [\,m\,]$ and all 
permutation $\mu$ on $[\,m\,]$. Since this is true for all permutations, it is also true if we select 
$\tilde{\mu}$ {\em randomly}, independently of $\{A_i\}$, with any distribution on the (finite) set of all possible permutations on $[\,m\,]$.

As $\{A_i\}$ are almost surely non-equal, and function $\psi$ has properties P1 and P2, it holds with probability one that 
\begin{equation}
\psi\big(\,A_{\sigma(1)}, \dots, A_{\sigma(m)}\,\big) \, = \;\psi\big(\,A_{\mu(1)}, \dots, A_{\mu(m)}\,\big),
\end{equation}
if and only if $\sigma(1) = \mu(1)$, where $\sigma, \mu$ are permutations on $[m]$. Hence, there are $m$ equivalence classes of permutations, denoted by $P_1, \dots, P_m$, each containing $(m-1)!$ permutations, with the (a.s.) property that permutations from the same class produce the same rank, while permutations from different classes produce different ranks. Therefore, each rank $k\in [\,m\,]$ is produced by exactly one class $P_i$, but naturally, the association of ranks and classes depends on the realization of the random elements $A_1, \dots, A_m$.

Now, let us fix a realization $a_1, \dots, a_m \in \BA$ in which the elements are pairwise different. Then, let us sample a permutation $\tilde{\mu}$ randomly, with {\em uniform} distribution on the set of all permutations. Since each equivalence class has the same number of elements, the probability that $\tilde{\mu} \in P_i$ is exactly $1/m$. As each $P_i$ yields a different rank, we have
\begin{equation}
\label{eq:unif-rank-fixed-realization}
\PP \big(\,\psi\big(\,a_{\tilde \mu(1)}, \dots, a_{\tilde \mu(m)}\,\big)\, = \, k\,\big) \; = \; \nicefrac{1}{m},
\end{equation}
for all rank $k\in [\,m\,]$ and {\em independently} of the realization $a_1, \dots, a_m$. Note that if we did not use a uniform distribution, then the resulting rank distribution would of course depend on the actual realization we are ranking. 

Because \eqref{eq:unif-rank-fixed-realization} is independent of the realization, the resulting discrete uniform distribution carries over to the case when $A_1,\dots, A_n$ are random, as they are (a.s.) pairwise different. 

This last step can be made more precise as follows. For simplicity, let us introduce the notations $a \doteq (a_1, \dots, a_m)$, $a_{\mu} \doteq (a_{\mu(1)}\, \dots, a_{\mu(m)})$, and similarly for $A$ and $A_{\mu}$. Then, let us introduce the indicator function of the rank being $k$,
\begin{equation}
\BI_k(a, \mu) \;\doteq\;
 \begin{cases}
    \,1, & \text{if } \psi(a_{\mu}) \,=\, k, \\
    \,0, & \text{otherwise},
  \end{cases}
\end{equation}
where $a$ and $\mu$ are deterministic. Then, let us define
\begin{equation}
i_k(a) \; \doteq \; \EE \big[\, \BI_k(a, \tilde{\mu}) \,\big],
\end{equation}
where $\tilde{\mu}$ is a {\em random} permutation selected uniformly from the set of all permutations on  $[\,m\,]$, and $a\in \BA^m$ is a constant. Note that $i_k(\cdot)$ is a deterministic function. Then, we have
\begin{equation}
i_k(a)\, =\, \PP \big(\,\psi(a_{\tilde{\mu}}) = \, k\,\big) \, = \, \nicefrac{1}{m},
\end{equation}
for all $a$ whose elements are pairwise different. Then, using the properties of (conditional) expectation, we have
\begin{align}
\PP \big(\,\psi(A)\, =\, k\,\big) &= \, \PP \big(\,\psi(A_{\tilde{\mu}}) = \, k\,\big)\, =\,\EE\big[ \, \BI_k(A, \tilde{\mu}) \, \big] \nonumber\\[1mm]
&= \,\EE\big[ \,\EE\big[\, \BI_k(A, \tilde{\mu})\mid A\,\big] \, \big]\,=\,\EE\big[ \,i_k(A)\,\big]\nonumber\\[1mm]
&= \, \EE\big[ \,\nicefrac{1}{m}\,\big] \, = \, \nicefrac{1}{m},
\end{align}
where we also used that the elements of $A$ are almost surely pairwise different. This concludes the proof of the claim.
\end{proof}

\subsection{Confidence Regions}
Inspired by FSID methods \cite{Algo2018, SPSPaper2ITA, KolumbanThesis2016},
the core idea of the proposed algorithms is to compare the original dataset with alternative samples which are randomly generated according to a given hypothesis. The comparison will be based on the rank of the original dataset among all the available samples, therefore, the ranking function is in the heart of all proposed algorithms. The differences between various algorithms primarily come from the various ways they rank. 

Lemma \ref{lemma:discrete-uniform-ranks} will be one of our main technical tools, however, it requires almost surely different elements, which is not guaranteed for $\{\CD_k(\theta)\}$. This will be resolved by {\em random tie-breaking}, similarly to the solution of \cite{SPSPaper2ITA}. To make this precise, consider a permutation $\pi$ of the set $\{0,\dots, m-1\}$, generated randomly with {\em uniform} distribution, and independently of $\{\CD_k(\theta)\}$. Then, obviously $\pi(0), \dots, \pi(m-1)$ are almost surely different, {\em exchangeable} random variables. 

We extend datasets $\{\CD_k(\theta)\}$ with $\{\pi(k)\}$. As a shorthand notation we introduce, for $k = 0,\dots, m-1$, the sample
\begin{equation}
\CD_k^{\pi}(\theta)\,\doteq\, \big(\CD_k(\theta),\pi(k)\big),
\end{equation}
which now takes values in\, $\BA \, \doteq\,\BS^n\times \{0,\dots, m-1\}$.

Given a ranking function $\psi$, defined on the codomain (range) of the extended datasets, and hyper-parameters $p, q  \in [\,m\,]$ with $p\, \leq\, q$, a {\em confidence region} can be defined by
\vspace{1mm}
\begin{equation}
\label{eq:confidence-region}
\Theta_{\varrho}^{\psi} \, \doteq \,\big\{ \, \theta \in \Theta :\,  p\, \leq\, \psi\big(\,\CD^{\pi}_0, \{ \CD^{\pi}_k(\theta) \}_{k \neq 0}\,\big)\, \leq\, q\, \big\},
\vspace{1mm}
\end{equation}
where $\varrho\, \doteq\, (m,p,q)$ denotes the applied hyper-parameters, with $m \geq 1$ being the total number of available samples, including the original one as well as the generated ones.

Our main abstract result about the {\em coverage probability} of the true parameter of such confidence regions is
\medskip
\begin{theorem}
\label{theorem:exact-confidence}
{\em 
We have for all ranking function $\psi$ and hyper-parameter $\varrho = (m,p,q)$ with 
integers $1\, \leq\, p\, \leq\, q\, \leq\, m$,}
\begin{equation}
\PP\big(\, \theta^* \in \Theta_{\varrho}^{\psi}  \, \big)\; = \; \frac{q-p+1}{m}.
\vspace{3mm}
\end{equation}
\end{theorem} 
\smallskip
\begin{proof}
First note that $\CD_0, \CD_1(\theta^*), \dots, \CD_{m-1}(\theta^*)$ are {conditionally i.i.d.}, given the inputs, $\{x_k\}$, therefore they are also {exchangeable}. As $\pi(0), \dots, \pi(m-1)$ are exchangeable, as well, and $\pi$ is generated independently of the datasets, we have that $\CD_0^{\pi}, \CD^{\pi}_1(\theta^*), \dots, \CD^{\pi}_{m-1}(\theta^*)$ are {\em exchangeable}, too, furthermore, they are almost surely pairwise different. 

Then, the theorem follows directly from Lemma \ref{lemma:discrete-uniform-ranks}, as the lemma implies that the rank of $D^{\pi}_0$ takes each value in $[\,m\,]$ with probability exactly $\nicefrac{1}{m}$, therefore, the probability that its rank is between $p$ and $q$ is exactly $(\,q-p+1\,)\,/\,m$.
\end{proof}
\medskip

Theorem \ref{theorem:exact-confidence} shows that the confidence regions constructed as \eqref{eq:confidence-region} have {\em exact} coverage probabilities, independently of the underlying probability distribution generating the (i.i.d.) data and for all ranking functions (satisfying P1 and P2). Observe that it is a {\em non-asymptotic} result, the exact coverage probability is valid irrespective of the sample size, $n$. Also note that the hyper-parameters are user-chosen, therefore, any (rational) confidence probability in $(0,1)$ can be achieved.

This theorem is very general and hence also allows some degenerate constructions, like the ones that do not depend on the data at all, only on the tie-breaking random permutation, $\pi$. Such regions are called {\em purely randomized}. In order to avoid such constructions, we should analyze other properties of the methods. Besides having guaranteed confidence, one of the most important properties an algorithm can have is (strong) consistency, namely, the property that, for any false parameter, 
as the sample size increases, eventually it will be excluded from the constructed confidence region (a.s.).

Formally, a method is {\em strongly consistent} if
\begin{equation}
\mathbb{P}\,\bigg(\,\bigcap_{k=1}^{\infty} \bigcup_{n=k}^{\infty} \left\{\,\theta \in \Theta_{\varrho, n}^{\psi} \,\right\} \bigg) \,=\, 0,
\label{eq:strongly-consistent-def}
\vspace{0mm}
\end{equation}
for all parameter $\theta\, \neq\, \theta^*$, $\theta \in \Theta$, where $\Theta_{\varrho, n}^{\psi}$ denotes the confidence region constructed based on a sample of size $n$. Obviously, purely randomized regions are not consistent.

\medskip
\section{Kernel-Based Constructions}
In this section we propose three kernel-based algorithms to construct confidence regions based on the resampling framework of Section \ref{section:resampling-framework}. We show that all of these methods have {\em exact} coverage probabilities and are {\em strongly consistent}.

\subsection{Algorithm I (Neighborhood Based)}
The main idea of Algorithm 1 is that we can estimate the regression function, $f_*$, based on the available (quantized) dataset, $\CD_0$, by the kNN ($k$-nearest neighbors) algorithm. We can similarly do so based on the alternative datasets, $\{\CD_k(\theta)\}_{k\neq 0}$. Then, we can compare the estimate based on $\CD_0$ to the ones coming from the alternative samples.

For Algorithm I we assume that $\BX \,\subseteq\, \RR^d$, $\BX$ is {\em compact}, the {\em support} of the (marginal) distribution of the inputs, $P_{\hspace{-0.7mm}\scriptscriptstyle\BX}$, is the whole $\BX$, furthermore, $P_{\hspace{-0.7mm}\scriptscriptstyle\BX}$ is {\em absolutely continuous}. 

Let us introduce functions, for $i = 0, \dots, m-1$, as
\vspace{-0.5mm}
\begin{equation}
f_{\theta, n}^{(i)}(x) \, \doteq \, \frac{1}{k_n}\, \sum_{j =1}^n \,y_{i,j}(\theta)\, \BI\hspace{0.3mm}\big(\,x_j \in N(x, n_k)\,\big),
\label{eq:knn-estimates}
\end{equation}
where $\BI$ is an indicator function (its value is $1$ if its argument is true, and $0$ otherwise), $N(x, n_k)$ denotes the $k_n$ closest neighbors of $x$ from 
$\{x_{j}\}_{j=1}^{n}$, and $k_n \leq n$ is a constant (window size), which
can depend on $n$. We use the standard Euclidean distance as a metric on $\BX$ (to define neighbors). Since the inputs, $\{x_j\}$, have a distribution that is absolutely continuous,
there is zero probability of ties in $N(x, n_k)$.

Given two square integrable functions, $f, g : \BX \to \RR$, let
\begin{equation}
\|\hspace{0.4mm} f - g\hspace{0.4mm}\|^2_2\, \doteq \int_{\BX} (f(x)-g(x))^2 dx.
\label{eq:L2-distance-unif}
\end{equation}

We will need the total (cumulative) distance of $f_{\theta, n}^{(i)}$ from all other functions, thus we introduce, for $i = 0, \dots, m-1,$
\begin{equation}
\label{eq:def-Z_i-knn}
Z_n^{(i)}(\theta) \, \doteq\, \sum_{j=0}^{m-1} \|\hspace{0.4mm} f_{\theta, n}^{(i)} - f_{\theta, n}^{(j)}\hspace{0.4mm}\|^2_2.
\end{equation}
Then, we can define the {\em rank} of $Z_n^{(0)}$ among $\{Z_n^{(i)}(\theta)\}$ as
\begin{equation}
\mathcal{R}_n(\theta)\; \doteq \; 1 \,+\, \sum_{i=1}^{m-1} \BI\hspace{0.3mm} \big(\,Z_n^{(0)} \prec_{\pi} Z_n^{(i)}(\theta)\,\big),
\label{eq:rank-knn}
\end{equation}
where $\BI$ is an indicator function, and binary relation ``$\prec_{\pi}$'' is the standard ``$<$'' with random tie-breaking. More precisely, as before, let $\pi$ be a random (uniformly chosen) permutation of the set $\{ 0, \dots, m-1 \}$. Then, given $m$ arbitrary real numbers, $Z_0, \dots, Z_{m-1}$, we can construct a strict total order, denoted by ``$\prec_{\pi}$'', by defining $Z_k \prec_{\pi} Z_j$ if and only if $Z_k < Z_j$ or it both holds that $Z_k = Z_j$ and $\pi(k) < \pi(j)$.

Therefore, in case of Algorithm I,  the ranking function is
\begin{equation}
\psi\big(\,\CD^{\pi}_0, \{ \CD^{\pi}_k(\theta) \}_{k \neq 0}\,\big)\, =\, \mathcal{R}_n(\theta).
\end{equation}
As we will see (cf.\ the proof of Theorem \ref{theorem:algorithm-I}), for any fixed false parameter, $Z_n^{(0)}(\theta)$ tends to have the largest rank, therefore, we fix $p = 1$ and only exclude parameters which lead to high ranks. That is, using \eqref{eq:confidence-region}, the confidence set is
\vspace{1mm}
\begin{equation}
\label{eq:conf-region-knn}
\Theta_{\varrho, n}^{(1)} \, \doteq \,\big\{ \, \theta \in \Theta :\,  \CR_n(\theta)\, \leq\, q\, \big\},	
\vspace{1mm}
\end{equation}
where $\varrho\, \doteq\, (\,m, q\,)$ again denotes the user-chosen hyper-parameters with
$1 \,\leq\, q\, \leq\, m$; we assume that $3\, \leq \, m$.

The main theoretical results can be summarized as
\medskip
\begin{theorem}
\label{theorem:algorithm-I}
{\em The coverage probability of the region is 
\begin{equation}
\PP\big(\,\theta^* \in \Theta_{\varrho,n }^{(1)}\,\big)\, = \, q\,/\,m,
\end{equation}
for any sample size $n$. Moreover, if $\{k_n\}$ are chosen such that $k_n \to \infty$ and $k_n/n \to 0$, as $n \to \infty$, then the confidence regions are strongly consistent, as defined by \eqref{eq:strongly-consistent-def}.}
\end{theorem}
\medskip
\begin{proof}
The exact confidence of the constructed regions immediately follows from Theorem \ref{theorem:exact-confidence}, as it is straightforward to check that the applied ranking satisfies P1 and P2.

In order to prove strong consistency, let us fix a false parameter $\theta \in \Theta$ with $\theta \neq \theta^*$. Since the parametrization is injective, we know that $f_{\theta} \neq f_*$ on a set of positive measure.

Under our assumptions we know that the kNN estimator \eqref{eq:knn-estimates} is strongly consistent \cite[Theorem 23.7]{gyorfi2002distribution}, that is
\begin{align}
\|\hspace{0.2mm} f_{\theta, n}^{(i)} -  f_{\theta}  \|^2_{\scriptscriptstyle P} \to 0& \quad\mbox{as}\quad n \to \infty,\\[1mm]
\|\hspace{0.2mm} f_{\theta, n}^{(0)}- f_* \|^2_{\scriptscriptstyle P} \to 0 & \quad\mbox{as}\quad n \to \infty,
\end{align}
almost surely, for $i=1,\dots, m-1$. Since the support of $P_{\hspace{-0.7mm}\scriptscriptstyle\BX}$ is $\BX$ and it is absolutely continuous, we have the same (a.s.) convergence properties if we use $\|\cdot\|_2^2$ instead of 
$\| \cdot  \|^2_{\scriptscriptstyle P}$. Now, let  $\kappa \,\doteq\, \| f_* - f_{\theta} \|^2_{2} > 0$, then  taking \eqref{eq:def-Z_i-knn} into account,
\begin{align}
Z_{n}^{(i)}(\theta) \to \kappa& \quad\mbox{as}\quad n \to \infty,\\[1mm]
Z_{n}^{(0)}(\theta) \to (m-1)\kappa& \quad\mbox{as}\quad n \to \infty,
\end{align}
almost surely, from which $Z_{n}^{(0)}(\theta)$ tends to take rank $m$ in the ordering (a.s.), as $n \to \infty$. Therefore, for any fixed $\theta\neq \theta^*$, asymptotically we (a.s.) have that $\mathcal{R}_{\infty}(\theta) = m$, which means that $\theta \neq \theta^*$ will be (a.s.) excluded from the confidence region (since $q < m$), as the sample size tends to infinity.
\end{proof}
\medskip

Regarding the computation aspects of Algorithm I note that $\{f_{\theta, n}^{(i)}\}$ can be calculated exactly based on the available data, as they are piece-wise constant functions. The distance $\|\hspace{0.4mm} f_{\theta, n}^{(i)} - f_{\theta, n}^{(j)}\hspace{0.4mm}\|^2_2$ can also be calculated from the available data. Nevertheless, one may use the Monte Carlo approximation\vspace{-0.5mm}
\begin{equation}
\|\hspace{0.4mm} f_{\theta, n}^{(i)} - f_{\theta, n}^{(j)}\hspace{0.4mm}\|^2_2 \; \approx\, \frac{1}{\ell_n}\,\sum_{k=1}^{\ell_n} (f_{\theta, n}^{(i)}(\bar x_k) - f_{\theta, n}^{(j)}(\bar x_k))^2,
\label{eq:emp-L2-distance}
\end{equation}
where $\ell_n$ is a constant and $\{\bar x_k\}$ are i.i.d.\ random variables having uniform distribution on $\BX$.
Note that 
we know from the {\em strong law of large numbers} (SLLN) that the sum in \eqref{eq:emp-L2-distance} almost surely converges to $\|\hspace{0.4mm} f_{\theta, n}^{(i)} - f_{\theta, n}^{(j)}\hspace{0.4mm}\|^2_2 $, as $\ell_n\to \infty$.

It is relatively easy to see that using the approximation in \eqref{eq:emp-L2-distance}, instead of \eqref{eq:L2-distance-unif}, does not affect the {\em exact} coverage probability of the algorithm. Moreover, if $\ell_n \to \infty$ as $n \to \infty$, then one can also show the strong consistency of the Monte Carlo approximated variant. Hence, the theoretical properties of Theorem \ref{theorem:algorithm-I} remain valid even under \eqref{eq:emp-L2-distance}, but the sizes of regions are of course affected by the approximation.

The kNN estimator, which is in the core of Algorithm I, is a simple kernel method that uses a variable bandwidth rectangular window. A natural generalization of this approach is to apply other kernels, such as Gaussian or Laplacian, for local averaging. Given any kernel $k(\cdot, \cdot)$, by interpreting it as a similarity measure, we can redefine functions $\{f_{\theta, n}^{(i)}\}$ as
\begin{equation}
f_{\theta, n}^{(i)}(x) \, \doteq \, \frac{1}{\sum_{l=1}^m k(x, x_l)} \sum_{j = 1}^n y_{i,j}(\theta)\, k(x, x_j),
\vspace{1mm}
\label{eq:kernelized-knn-estimates}
\end{equation}
which leads to alternative confidence region constructions. 

These variants typically also 
build confidence regions with {\em exact} coverage probabilities. Moreover, as a wide variety of such kernel estimates are strongly consistent, under some technical conditions \cite{gyorfi2002distribution}, and the generalized Algorithm I inherits these properties,
the resulting 
confidence sets are also {\em strongly consistent}.  The corresponding coverage and consistency theorems could be proved analogously to Theorem \ref{theorem:algorithm-I}.
\medskip
\subsection{Algorithm II (Embedding Based)}
The core idea of Algorithm II is to embed the distribution of the original sample and that of the alternative ones in an RKHS using a characteristic kernel. If the underlying distributions are different, then the original dataset results in a different element than the one the alternative datasets are being mapped to, which can be detected statistically.

Assume $\CH$ is a {separable} RKHS  containing $\BS \to \RR$ type functions with a 
{\em characteristic}, bounded, and translation-invariant kernel $k(\cdot, \cdot)$. 
If $\BX = \RR^d$, then $\BS = \RR^{d}\times\{+1,-1\}$, and 
we can use, for example, the Gaussian, the Laplacian, or the Poisson kernel,
which are all  characteristic \cite{muandet2017kernel}. 

Let us introduce the following kernel mean embeddings
\begin{align}
h_*(\cdot) &\,\doteq\, \EE \big[\,k(\cdot, S_*)\,\big],\\[2mm]
h_{\theta}(\cdot)&\, \doteq\, \EE \big[\,k(\cdot, S_{\theta})\,\big],
\end{align}
where $S_*$ and $S_{\theta}$ are a random elements from $\BS$; Variable $S_*$ has the ``true'' distribution of the observations, while $S_{\theta}$ has a distribution where the output, $Y$, is generated according to the conditional probability \eqref{eq:param-cond-prob}, parametrized by $\theta$, while the marginal distribution of the input, $X$, remains the same.

Since the kernel is bounded, $\EE \big[\,\sqrt{k(S_{\theta}, S_{\theta})}\,\big] < \infty$, for all $\theta$, which ensures that 
$\{h_{\theta}\}$ exist and belong to $\CH$ \cite{muandet2017kernel}.

Because the kernel is characteristic, we know that $h_{\theta} = h_*$ if and only if $\theta = \theta^*$. Now, let us introduce the following empirical versions of the embedded distributions, 
\begin{equation}
h_{\theta,n}^{(i)}(\cdot) \,\doteq \, \frac{1}{n}\sum_{j=1}^{n}\, k(\cdot, s_{i,j}(\theta)),
\end{equation}
for $i=0, \dots, m-1$, where $s_{i,j}(\theta) \doteq (x_j, y_{i,j}(\theta))$; and recall that for $i=0$ (original sample), we have $y_{i,j}(\theta) = y_j$.  In other words, $s_{i,j}(\theta)$ has the same distribution of $S_{\theta}$ for $i\neq0$ and its distribution is the same as that of $S_*$ for $i = 0$.

Let  $C_k$ be a constant that satisfies $|\,k(x,y)\,| \leq C_k$ for all $x,y$. Then, obviously $|\,h_{\theta}(x)\,|\leq C_k$ for all $x$, as well.
Now, applying 
the reproducing property, we have the bound
\begin{align}
\Var(k(\cdot, S)) &\,=\, \EE \big[\, \|\, k(\cdot, S) - h(\cdot)\, \|^2_{\CH}\, \big]\nonumber \\[1mm]
&\,\leq \, \EE \big[\, \|\, k(\cdot, S)\,\|^2_{\CH} \, \big] +\, \EE \big[\, \|\,  h(\cdot)\, \|^2_{\CH}\, \big]\nonumber \\[1mm]
&\,\;\;\;\;+\, 2  \hspace{0.5mm}  \EE \big[\,  | \left<\,k(\cdot, S),  h(\cdot)\,\right>_{\CH} | \, \big]\nonumber\\[1mm]
&\,\leq \, \EE \big[\, \|\, k(\cdot, S)\,\|^2_{\CH}\, \big] + \|\,  h(\cdot)\, \|^2_{\CH} + 2  \hspace{0.5mm}\EE \big[\,  |\,  h(S)\, | \, \big]\nonumber\\[1mm]
&\,=\, \EE \big[\, \left<\, k(\cdot, S), k(\cdot, S)\, \right>_{\CH} \,\big]+  \|\,  h(\cdot)\, \|^2_{\CH} + 2\hspace{0.5mm} C_k\nonumber \\[1mm]
&\,=\,\EE \big[\,k(S, S) \, \big] + \|\,  h(\cdot)\, \|^2_{\CH} + 2\hspace{0.5mm} C_k\nonumber \\[1mm]
&\,= \,3\hspace{0.5mm} C_k + \|\,  h(\cdot)\, \|^2_{\CH}\, <\, \infty,
\label{eq:variance-bound}
\end{align}
where $S$ is either $S_*$ or $S_{\theta}$, and $h\, \doteq\, \EE \big[\, k(\cdot, S)\, \big]$.

Then, we know from the SLLN for Hilbert space valued elements that $\|\,h_{\theta,n}^{(i)} - h_{\theta}\,\|_{\CH}\to 0$ (a.s.), as $n \to \infty$, for $i \neq 0$, additionally, $\|\,h_{\theta,n}^{(0)} - h_{*}\,\|_{\CH}\to 0$ (a.s.), as $n \to \infty$.

Now, we can define the $\{Z_n^{(i)}\hspace{-0.4mm}(\theta)\}$ variables similarly to \eqref{eq:def-Z_i-knn}, but using the squared distances $\|\,h_{\theta,n}^{(i)} - h_{\theta,n}^{(j)}\,\|^2_{\CH}$ instead of $ \|\hspace{0.4mm} f_{\theta, n}^{(i)} - f_{\theta, n}^{(j)}\hspace{0.4mm}\|^2_2$,
and construct the confidence set as \eqref{eq:conf-region-knn}.

\medskip
\begin{theorem}
\label{theorem:algorithm-II}
{\em The confidence regions of Algorithm II have
\begin{equation}
\PP\big(\,\theta^* \in \Theta_{\varrho,n }^{(2)}\,\big)\, = \, q\,/\,m,
\end{equation}
for any sample size $n$; and they are strongly consistent.}
\end{theorem}
\medskip
\begin{proof}
The exact confidence again follows from Theorem \ref{theorem:exact-confidence} by noting that the ranking satisfies P1 and P2.

The proof of consistency follows the ideas of the proof of Theorem \ref{theorem:algorithm-I}. Namely, let us fix a false parameter $\theta \in \Theta$ with $\theta \neq \theta^*$. Since the parametrization is injective, we know that $\CD_0$ and $\{
\CD_i(\theta)\}_{i\neq0}$ have different distributions. As the kernel is characteristic, we know that the RKHS embedded distributions $h_*(\cdot)$ and $h_{\theta}(\cdot)$ are different. We then apply the SLLN for Hilbert space valued elements \cite{taylor1978stochastic} and  use the construction of the $\{Z_n^{(i)}\}$ variables to get the limits
\begin{align}
Z_{n}^{(i)}(\theta) \to \kappa& \quad\mbox{as}\quad n \to \infty,\\[1mm]
Z_{n}^{(0)}(\theta) \to (m-1)\kappa& \quad\mbox{as}\quad n \to \infty,
\end{align}
for $i \neq 0$, almost surely, where $\kappa \,\doteq\, \| h_* - h_{\theta} \|_{\CH} > 0$. Thus, $Z_{n}^{(0)}(\theta)$ again tends to take rank $m$ (a.s.), as $n \to \infty$, which leads to the (a.s.) asymptotic exclusion of the false parameter $\theta \neq \theta^*$ (for more details, see the proof of Theorem \ref{theorem:algorithm-I}).
\end{proof}
\medskip
The squared distance of the empirical versions of the embeddings $\|\,h_{\theta,n}^{(i)} - ,h_{\theta,n}^{(j)} \,\|^2_{\CH}$ can be computed by applying the reproducing property of the kernel
and the Gram matrix of the sample $s_{i,1}(\theta), \dots, s_{i,n}(\theta), s_{j,1}(\theta), \dots, s_{j,n}(\theta)$.

Algorithm II has a nice theoretical interpretation as comparing embedded distributions in an RKHS. 
However, as the Gram matrices required to compute the $\{Z_n^{(i)}\hspace{-0.4mm}(\theta)\}$ variables
depend on $\theta$, this method has a large computational burden, hence the importance of Algorithm II is mainly theoretical. Nevertheless, motivated by its ideas, in the next section 
we suggest a computationally much lighter algorithm.

\medskip
\subsection{Algorithm III (Discrepancy Based)}
Algorithm III follows the intuitions behind Algorithm II, but ensures that we can work with the same Gram matrix for all $\theta$. Moreover, it has a simpler construction for $\{Z_n^{(i)}\hspace{-0.4mm}(\theta)\}$, which also makes it computationally more appealing.

For Algorithm III we assume that $\CH$ is a {separable} RKHS containing $\BX \to \RR$ functions with a {\em universal}, bounded, and translation-invariant kernel $k(\cdot, \cdot)$. We assume that $\BX$ is a {\em compact} metric space, hence, $k(\cdot, \cdot)$ is also {\em characteristic} \cite{muandet2017kernel}. Finally, we assume that each $f \in \CF$ is {\em continuous}.

Let us introduce the notation $\varepsilon_{i,j}(\theta)\, \doteq\, y_{i,j}(\theta) - f_{\theta}(x_j)$, for $i= 0, \dots, m-1$ and $j = 1, \dots, n$. Note that if $i \neq 0$, 
$\varepsilon_{i,j}(\theta)$ has zero mean for all $j$, as 
$f_{\theta}(x_j) =\, \EE_{\theta} \big[\, y_{i,j}(\theta) \,|\,x_j\,\big] $.

The fundamental objects of Algorithm III are
\vspace{-0.5mm}
\begin{equation}
Z_n^{(i)}\hspace{-0.2mm}(\theta)\, \doteq \, \bigg\|\, \frac{1}{n} \sum_{j=1}^n \varepsilon_{i,j}(\theta) \hspace{0.5mm}k(\cdot, x_j)\,\bigg\|^2_{\CH},
\vspace{-1mm}
\label{eq:Z_i-Alg-III}
\end{equation}
for $i= 0, \dots, m-1$. Observe that $Z_n^{(i)}\hspace{-0.2mm}(\theta)$ can be easily computed using the Gram matrix $K_{i,j} \,\doteq\, k(x_i, x_j)$, as
\begin{equation}
Z_n^{(i)}(\theta)\, = \, \frac{1}{n^2} \, \varepsilon\tr_i(\theta) \,K\, \varepsilon_i(\theta),
\end{equation}
using the notation $\varepsilon_i(\theta) \,\doteq\, (\varepsilon_{i,1}(\theta), \dots, \varepsilon_{i,n}(\theta))\tr$.

From this point, we follow the construction of Algorithms I and II, namely, we define the ranking function as \eqref{eq:rank-knn}, and the confidence region as \eqref{eq:conf-region-knn}, but naturally we apply our new functions \eqref{eq:Z_i-Alg-III} as the definition of the $\{Z_n^{(i)}\hspace{-0.4mm}(\theta)\}$ variables.

\medskip
\begin{theorem}
\label{theorem:algorithm-III}
{\em The confidence regions of Algorithm III have
\begin{equation}
\PP\big(\,\theta^* \in \Theta_{\varrho,n }^{(3)}\,\big)\, = \, q\,/\,m,
\end{equation}
for any sample size $n$; and they are strongly consistent.}
\end{theorem}
\medskip
\begin{proof}
The exact confidence follows from Theorem \ref{theorem:exact-confidence}.

For the proof of strong consistency, let us fix $\theta \neq \theta^*$ and an $i \neq 0$. To simplify the notations, introduce $e_j \doteq \varepsilon_{i,j}(\theta)$ and $\bar y_j \doteq y_{i,j}(\theta)$. We first show that $e_j k(\cdot, x_j)$ has zero mean 
\begin{align}
\hspace*{-2mm}\EE\big[\, e_j k(\cdot, x_j) \, \big]& =\, \EE\big[\, \EE [\,e_j k(\cdot, x_j) \mid x_j\, ] \, \big]\nonumber \\[1mm]
&=\,  \EE\big[\, \EE [\,(\bar y_j - f_{\theta}(x_j)) k(\cdot, x_j) \mid x_j\, ] \, \big]\nonumber\\[1mm]
&=\,  \EE\big[\, \EE [\,\bar y_j  \mid x_j\, ]k(\cdot, x_j) - f_{\theta}(x_j) k(\cdot, x_j)\, \big]\nonumber\\[1mm]
&=\,  \EE\big[\, (f_{\theta}(x_j) - f_{\theta}(x_j)) k(\cdot, x_j)\, \big] \,=\, 0.
\label{eq:ek-zero-mean}
\end{align}
About the variance of $e_j k(\cdot, x_j)$, observe that
\begin{equation}
\Var\big(\,e_j k(\cdot, x_j)\,\big) = \,\EE \big[\, \| \,  e_j k(\cdot, x_j)  \, \|^2_{\CH}\,\big]\, \leq \, 4\, B_k,
\end{equation}
where $B_k \doteq k(x,x)$, for any $x$ since the kernel is translation-invariant; also note that $\| k(\cdot, x)\|_{\CH}^2 = k(x,x)$, for any $x \in \BX$, because of the reproducing property of the kernel.

Therefore, we can apply the Hilbert space valued SLLN to conclude that $Z_{n}^{(i)}(\theta) \to 0$ (a.s.), as $n \to \infty$, for all $i \neq 0$.

Now, let $e^*_j \doteq \varepsilon_{0,j}(\theta) = y_j - f_{\theta}(x_j)$.
We will prove that the mean of $e^*_j k(\cdot, x_j)$ is not zero. We can again show
\begin{equation}
\EE\big[\, e_j k(\cdot, x_j) \, \big] =\, \EE\big[\, (f_*(x_j) - f_{\theta}(x_j)) k(\cdot, x_j)\, \big],
\end{equation}
using similar steps as in \eqref{eq:ek-zero-mean}, except in the last one, where in our case we have $\EE [\, y_j  \,|\, x_j\, ] = f_*(x_j)$. We will argue that the term $\EE\big[\, (f_*(x_j) - f_{\theta}(x_j)) k(\cdot, x_j)\, \big]$ cannot be zero.

Let us introduce $f_0 \doteq f_* - f_{\theta}$, and assume {\em indirectly} that $\EE\big[\, f_0(x_j)\, k(\cdot, x_j)\, \big]$ is the zero function. Then, for all $x$, $\left< f_0, k(x, \cdot) \right>_{\hspace{-0.5mm}\scriptscriptstyle P}\! \doteq \EE\big[\, f_0(x_j) k(x, x_j)\, \big] =\, 0$ (note that an RKHS is a space of functions and not that of equivalence classes of functions). Since the kernel is universal, $\BX$ is compact, and $f_0$ is continuous, we know that for all $\varepsilon >0$, there exists an $\hat{f} \in \CH$, such that $\| \,\hat{f} - f_0 \,\|_{\infty} < \varepsilon$. Then, clearly
\begin{align}
\int_{\BX} (\hat{f} - f_0)^2 P_{\hspace{-0.7mm}\scriptscriptstyle\BX}(dx) &\,\leq\, \int_{\BX} \| \hspace{0.5mm}\hat{f} - f_0 \hspace{0.5mm}\|_{\infty}^2\,P_{\hspace{-0.7mm}\scriptscriptstyle\BX}(dx) \nonumber\\[1mm]
&\,< \,\int_{\BX} \varepsilon^2 P_{\hspace{-0.7mm}\scriptscriptstyle\BX}(dx) \,=\, \varepsilon^2
\end{align}
since $P_{\hspace{-0.7mm}\scriptscriptstyle\BX}$ is a probability measure on $\BX$.
Hence, for all $\varepsilon > 0$,
\begin{equation}
\| \hspace{0.5mm}\hat{f} - f_0 \hspace{0.5mm}\|^2_{\scriptscriptstyle P} \, = \, \|\hspace{0.5mm}\hat{f}\hspace{0.5mm}\|^2_{\scriptscriptstyle P} + \|\hspace{0.5mm}f_0\hspace{0.5mm}\|^2_{\scriptscriptstyle P} - 2\hspace{0.5mm} \big<\hspace{0.5mm} f_0, \hat{f}\hspace{0.5mm} \big>_{\hspace{-0.5mm}\scriptscriptstyle P}\, <\, \varepsilon^2.
\label{eq:norm-decomposition}
\end{equation}

Since $k(\cdot, \cdot)$ is the kernel of the RKHS, we can write $\hat{f}$ as\!\!
\vspace{-0.7mm}
\begin{equation}
\hat{f}(\cdot)\, = \, \sum_{k=1}^{\infty} \alpha_k k(\cdot, \bar{x}_k),
\end{equation}
for some points $\{ \bar{x}_k\}$. Since for all $x$, $\left< f_0, k(x, \cdot) \right>_{\hspace{-0.5mm}\scriptscriptstyle P} = 0$, 
\vspace{-1mm}
\begin{align}
\big<\hspace{0.5mm} f_0, \hat{f}\hspace{0.5mm} \big>_{\hspace{-0.5mm}\scriptscriptstyle P} &=\, \int_{\BX} \,\sum_{k=1}^{\infty} \alpha_k k(x, \bar{x}_k) f_0(x) P_{\hspace{-0.7mm}\scriptscriptstyle\BX}(dx) \nonumber \\[1mm]
& =\, \sum_{k=1}^{\infty} \alpha_k \!\int_{\BX} k(x, \bar{x}_k) f_0(x) P_{\hspace{-0.7mm}\scriptscriptstyle\BX}(dx)  \nonumber\\[1mm]
& =\, \sum_{k=1}^{\infty} \alpha_k \left< f_0, k(\bar{x}_k, \cdot) \right>_{\hspace{-0.5mm}\scriptscriptstyle P} \,=\,0.
\label{eq:f0-hatf-orthog}
\end{align}
where we have applied Fubini's theorem \cite{kallenberg2002foundations} to exchange the two integrals (one of which is a sum). Regarding the applicability of Fubini's theorem note that both integrals are w.r.t.\ a finite measure, and the functions are bounded.

Then, combining  \eqref{eq:norm-decomposition} and \eqref{eq:f0-hatf-orthog} we get that for all $\varepsilon > 0$,
\begin{equation}
 \|\hspace{0.5mm}f_0\hspace{0.5mm}\|^2_{\scriptscriptstyle P} \, \leq \,  \, \|\hspace{0.5mm}\hat{f}\hspace{0.5mm}\|^2_{\scriptscriptstyle P} + \|\hspace{0.5mm}f_0\hspace{0.5mm}\|^2_{\scriptscriptstyle P} \,<\, \varepsilon^2,
\end{equation}
which implies that $\|\hspace{0.5mm}f_0\hspace{0.5mm}\|^2_{\scriptscriptstyle P} = 0$. On the other hand, we know from \eqref{eq:L2-distance-PX} that this norm cannot be zero if $\theta \neq \theta^*$. Therefore, we have reached a contradiction, hence $\EE\big[\, (f_*(x_j) - f_{\theta}(x_j)) k(\cdot, x_j)\, \big]$ cannot be the zero element of the RKHS.

We can use a similar argument to \eqref{eq:variance-bound} to show that $\Var(e^*_j k(\cdot, x_j))$ is bounded, also using that $\{e^*_j\}$ are bounded. Then, applying the Hilbert space variant of SLLN  \cite{taylor1978stochastic}, 
\begin{equation}
\frac{1}{n} \sum_{j=1}^{n}e^*_j k(\cdot, x_j) \,\to\, h_0\, \neq\, 0, \quad\mbox{as}\quad{n\,\to\,\infty},
\end{equation}
almost surely. Therefore, summarizing our results, we have
\begin{align}
Z_{n}^{(i)}(\theta) \to 0& \quad\mbox{as}\quad n \to \infty,\\[1mm]
Z_{n}^{(0)}(\theta) \to \|h_0\|^2_{\CH}& \quad\mbox{as}\quad n \to \infty,
\end{align}
for $i \neq 0$, almost surely, where $\|h_0\|^2_{\CH} > 0$. Thus, $Z_{n}^{(0)}(\theta)$ again tends to take rank $m$ (a.s.), as $n \to \infty$, which leads to the (a.s.) asymptotic exclusion of the parameter $\theta \neq \theta^*$.
\end{proof}
\vspace*{-5mm}
\medskip

\medskip
\section{Numerical Experiments}
\label{section:numerical-experiments}
Numerical experiments were carried out to demonstrate the proposed algorithms. In the presented test scenario the joint probability distribution of the data was assumed to be the mixture of two Laplace distributions with different locations, $\mu_1, \mu_2$, but with the same scale $\lambda$. It was assumed that with probability $p$ we observe the ``$+1$'' class, and with $1-p$ we see an element of the ``$-1$'' class. Selecting $p$, $\mu_1$, $\mu_2$ and $\lambda$ induces a regression function, e.g., see \eqref{eq:mixing-distributions}.

\begin{figure}[t]
	\vspace*{1mm}
    \centering
 	\subfigure[Algorithm I (kNN) Ranks]{\label{fig:LS-SVC}\includegraphics[height=34mm]{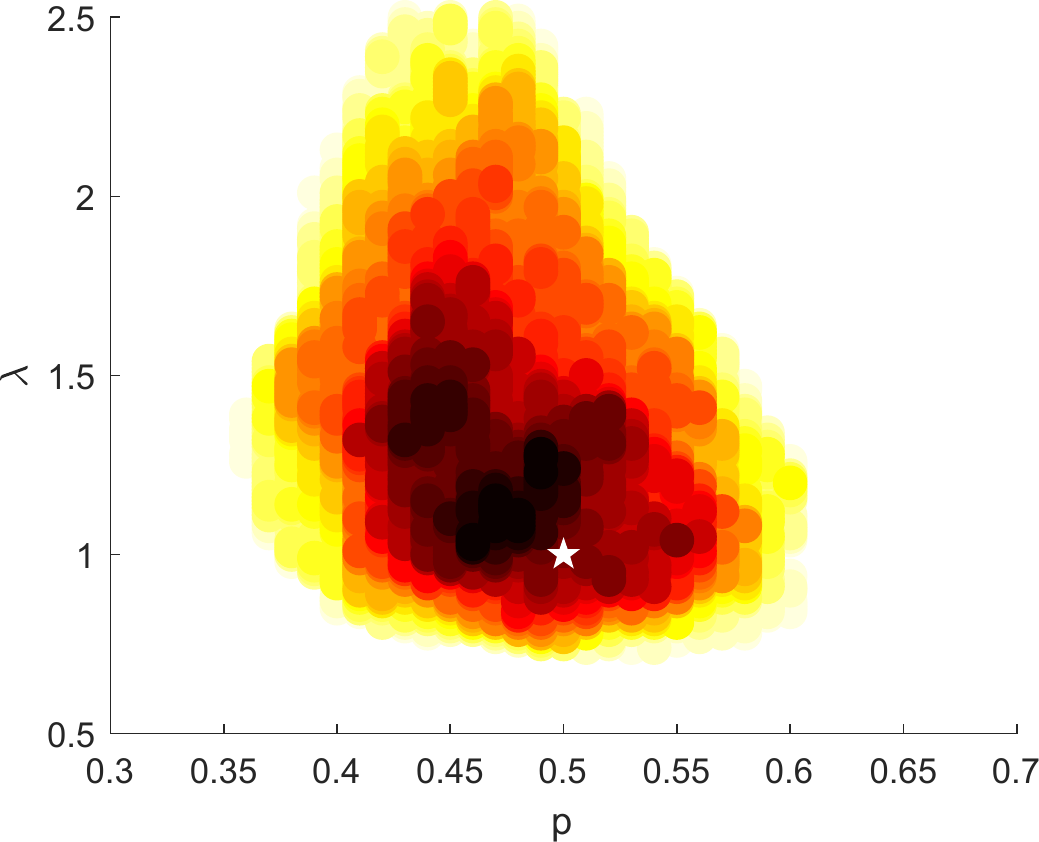}}
 	\subfigure[Algorithm I (Gauss) Ranks]{\label{fig:DPUniform}\includegraphics[height=34mm]{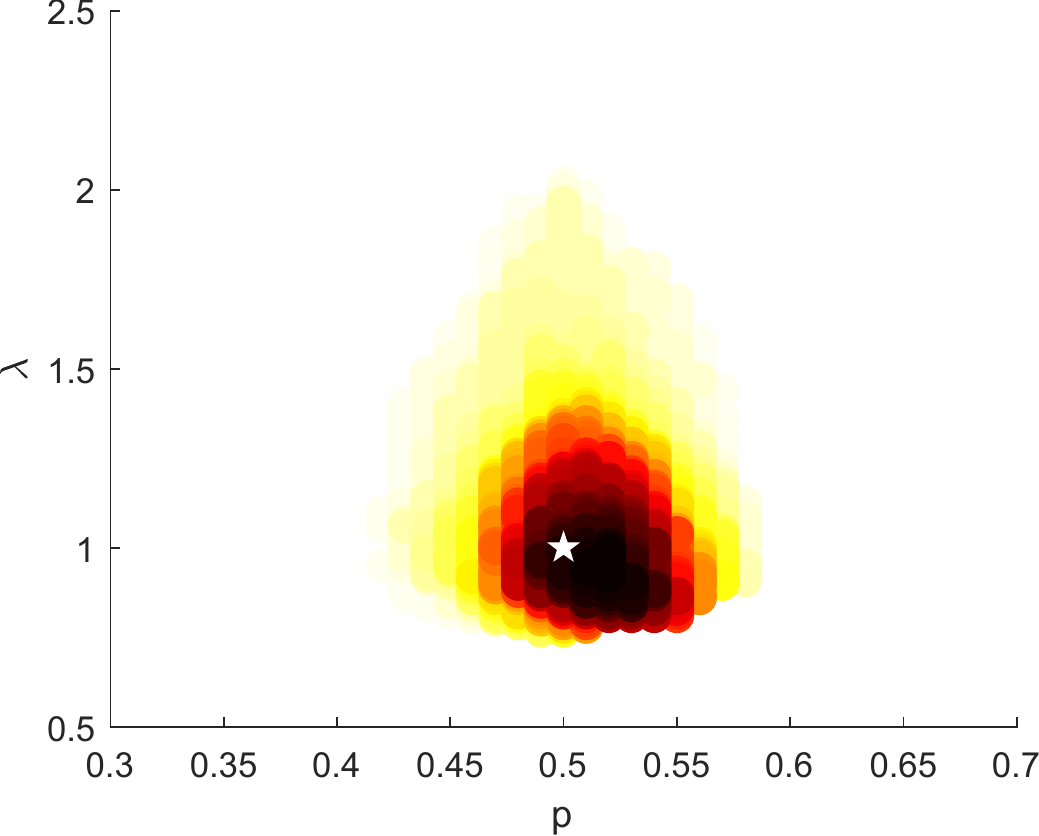}}\vspace{4mm}
	\subfigure[Algorithm III (Gauss) Ranks]{\label{fig:Assymptotic}\includegraphics[height=34mm]{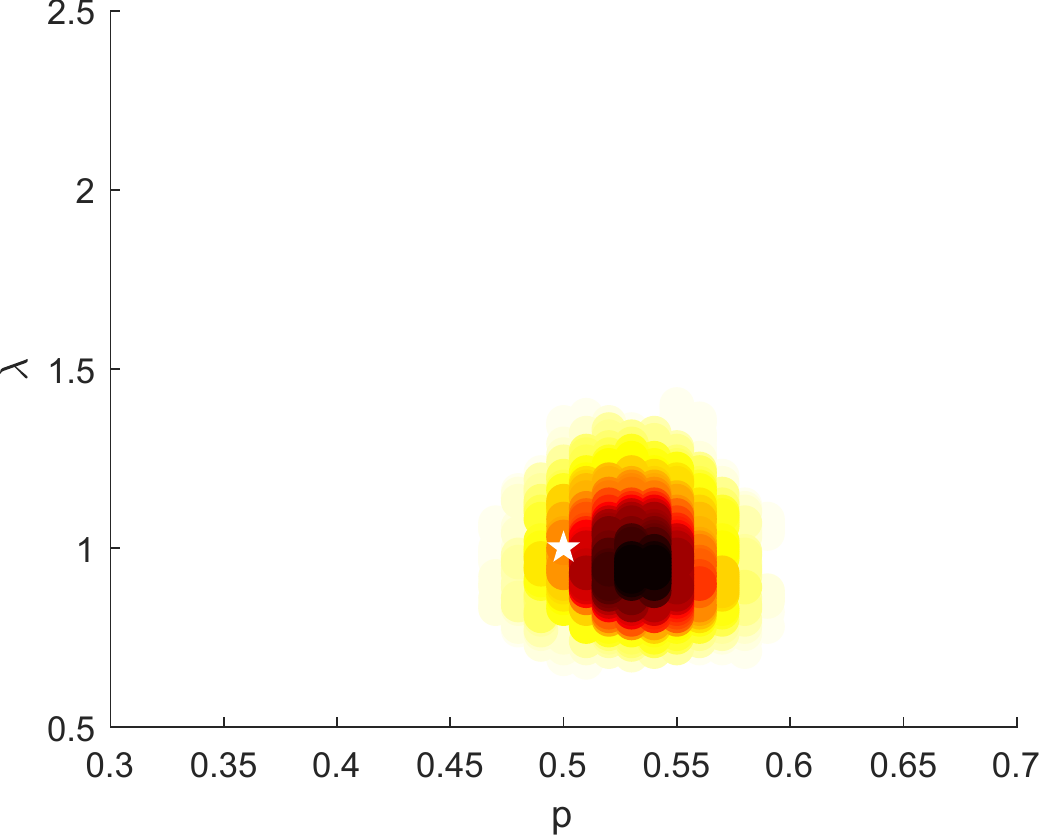}} 	
	\subfigure[$90\,\%$ Confidence Sets]{\label{fig:Assymptotic}\includegraphics[height=34mm]{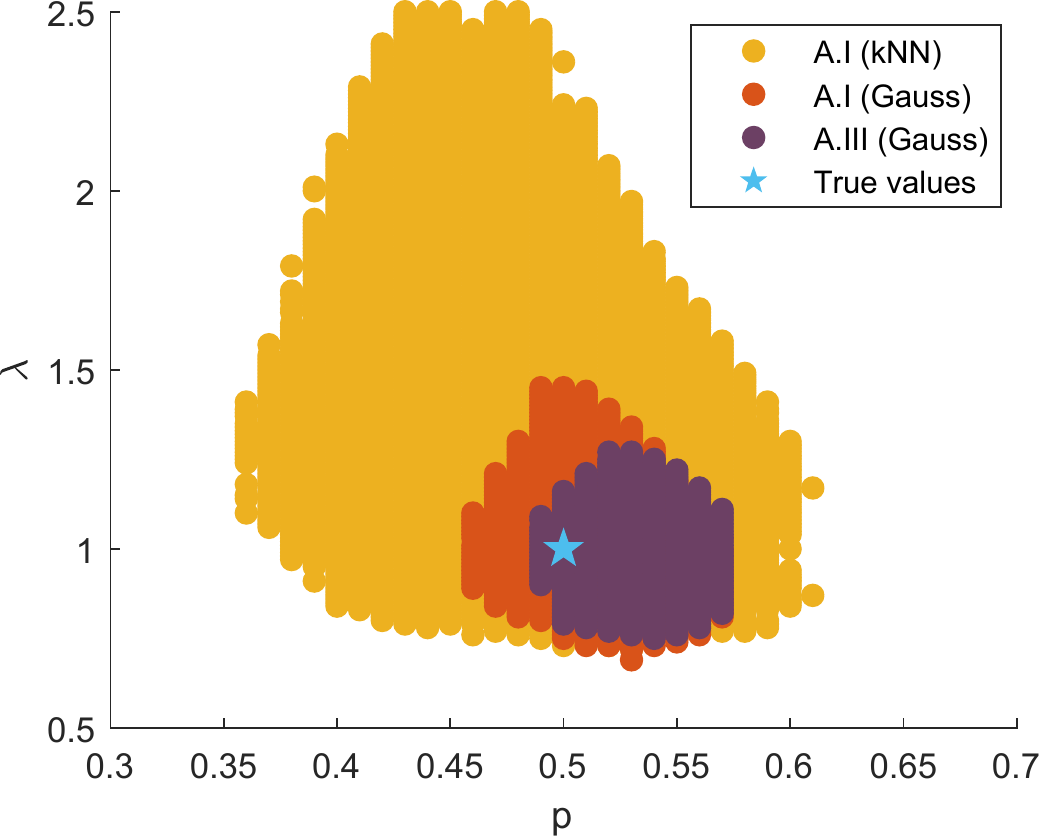}} 	
    \caption{The ranks (of the original sample) and (exact) confidence regions produced by the algorithms for various parameters. The model was assumed to be a mixture of Laplace distributions. The mixing probability $p = \nicefrac{1}{2}$ ($x$-axis) and the common scale parameter $\lambda = 1$ ($y$-axis) were estimated from a sample of size $n=500$. The ``$\star$'' denotes the true parameter.}
\label{fig:experiments}
\vspace*{-3mm}
\end{figure}

During the experiments the confidence regions were built for parameters $p$ and $\lambda$, while the location parameters were fixed, $\mu_1 = 1$ and $\mu_2 = -1$, to allow two dimensional figures. Figure \ref{fig:experiments} demonstrates the obtained ranks $\{\CR_n(\theta)\}$ for various $\theta = (\hspace{0.4mm}p, \lambda\hspace{0.4mm})$ using Algorithm I with the  kNN approach (a), Algorithm I with a Gaussian kernel (b) and Algorithm III with a Gaussian kernel (c). For the Gaussian kernel we choose $\sigma = \nicefrac{1}{2}$. On parts (a), (b) and (c) darker colors indicate smaller ranks, hence, the darker the color is, the more likely the parameter is included in a confidence region. The three corresponding $90\,\%$ (exact) confidence regions are also demonstrated by part (d). The true parameters were $p = \nicefrac{1}{2}$ ($x$-axis) and $\lambda = 1$ ($y$-axis). The sample size was $n = 500$ and $m = 50$ (original and alternative) samples were generated.  The regions were evaluated on a grid.

It can be observed that Algorithm III produced the most concentrated rank clusters and provided the smallest confidence region.
The extended version \eqref{eq:kernelized-knn-estimates} of Algorithm I, with a Gaussian kernel, produced comparable results, while the kNN version was the worst in this case. Nevertheless, it still has computational advantages which may make it attractive.

Note that in this special example it is possible to construct individual confidence regions for parameters $p$ and $\lambda$ based on standard results. One can use, for example, Hoeffding's inequality \cite{kallenberg2002foundations} to get confidence intervals for probability $p$, and $\lambda$ can be estimated based on the fact that the variance of the observations, for both classes, is $2 \lambda^2$. Nevertheless, such approaches need the {\em specific interpretations} of the parameters: on how they influence the observations. Furthermore, even in this very special case it is not obvious how to construct a {\em joint} confidence region for the $(\hspace{0.4mm}p, \lambda\hspace{0.4mm})$ pair. Simply intersecting the two confidence tubes (i.e., if we extend the confidence intervals for $p$ and $\lambda$ to $\RR^2$, then they define two infinite ``stripes'', a vertical and a horizontal one) produces a set with a lower confidence than that of the 
original sets, and hence it ultimately leads to {\em conservative} confidence regions.

On the other hand, the suggested three algorithms do not presuppose any interpretation of the tested parameters, apart from the fact that they determine a regression function. They do not need a fully parametrized joint distribution, indeed, the regression function is compatible with infinitely many 
joint  distributions having widely different (marginal) input  distributions. Furthermore, if $\theta \in \RR^d$, then the algorithms automatically build {\em joint} and {\em non-conservative} confidence sets. Hence, another advantage of the presented framework, apart from its strong theoretical guarantees, is its {\em flexibility}.

%
%

                                  %
                                  %
                                  %
                                  %
                                  %

%

\medskip
\section{Conclusions}

In this paper we addressed the problem of building {\em non-asymptotic} confidence regions for the {\em regression} function of binary classification, which is a key object defined as the conditional expectation of the class labels given the inputs.

The main idea was to test candidate models by generating {\em alternative samples} based on them, and then computing the performance of a kernel-based algorithm on all samples. If the candidate model is wrong, then the algorithm behave differently on the alternatively generated samples than on the original one, which can be detected statistically by {\em ranking}.

Three constructions were proposed and it was proved that all of them build confidence regions with {\em exact} coverage probabilities, for any sample size, and are {\em strongly consistent}. 

\startsquarepar
The proposed framework is semi-parametric, because the regression function does not determine the (joint) probability 
\stopsquarepar
\newpage
\noindent
distribution of the data, it does not contain information about the (marginal) distribution of the inputs (and that is why only the outputs are resampled in the alternative datasets). 

Moreover, the algorithms only {\em indirectly} depend on the given family of candidate 
functions, namely, their inputs are just the original sample and several alternative samples generated based on the tested 
function. Consequently, the family of regression functions
can be {\em arbitrary}. 
It could even be the set of all possible regression functions which satisfy \eqref{eq:L2-distance-PX} and the theoretical results are still valid. If we work with an infinite dimensional class of functions, then the confidence regions cannot be explicitly constructed in practice. Nevertheless, it is still possible to {\em test} any candidate regression function to check whether it is included in a confidence set, or in other words, to quantify its uncertainty by computing how compatible it is with the available observations.

\bigskip
\bibliographystyle{ieeetr}
\bibliography{kernel-classification} 

\end{document}